\documentclass[letterpaper, 10 pt, journal, twoside]{ieeetran}
\usepackage[letterpaper, left=48pt, right=48pt, bottom=46pt, top=57pt]{geometry}
\usepackage{graphicx}
\usepackage{amsmath} 
\usepackage{amssymb}  
\usepackage{mathtools}
\usepackage{bbm}
\usepackage{dsfont}
\usepackage{multirow}
\usepackage[dvipsnames]{xcolor}
\usepackage{subfig}
\usepackage[noadjust]{cite}
\usepackage{enumerate}
\usepackage{booktabs}
\usepackage{hyperref}
\usepackage{algpseudocode}
\usepackage{algorithm}
\usepackage{caption}
\begin{document}
\title{Learning Sparse Interaction Graphs of Partially \\ Detected Pedestrians for Trajectory Prediction}
\author{Zhe Huang\textsuperscript{1}, Ruohua Li\textsuperscript{2}, Kazuki Shin\textsuperscript{1}, and Katherine Driggs-Campbell\textsuperscript{1}
\thanks{Manuscript received: September 9, 2021; Revised November 25, 2021; Accepted December 14, 2021.}
\thanks{This paper was recommended for publication by Editor Jee-Hwan Ryu upon evaluation of the Associate Editor and Reviewers' comments. \textit{(Corresponding author:
Zhe Huang.)}
}
\thanks{$^1$ Z. Huang, K. Shin, and K. Driggs-Campbell are with the Department of Electrical and Computer Engineering at the University of Illinois at Urbana-Champaign, Urbana, IL 61801 USA (e-mail: zheh4@illinois.edu; kazukis2@illinois.edu; krdc@illinois.edu). \newline \indent $^2$ R. Li is with the Department of Electrical Engineering and Computer Science at the University of Michigan, Ann Arbor, MI 48109 USA (e-mail: ruohuali@umich.edu).}

\thanks{Digital Object Identifier (DOI): see top of this page.}
}

\markboth{IEEE Robotics and Automation Letters. Preprint Version. Accepted December, 2021}
{Huang \MakeLowercase{\textit{et al.}}: Learning Sparse Interaction Graphs of Partially Detected Pedestrians for Trajectory Prediction}



\maketitle
\begin{abstract}
Multi-pedestrian trajectory prediction is an indispensable element of autonomous systems that safely interact with crowds in unstructured environments. Many recent efforts in trajectory prediction algorithms have focused on understanding social norms behind pedestrian motions. Yet we observe these works usually hold two assumptions, which prevent them from being smoothly applied to robot applications: (1) positions of all pedestrians are consistently tracked, and (2) the target agent pays attention to all pedestrians in the scene. The first assumption leads to biased interaction modeling with incomplete pedestrian data. The second assumption introduces aggregation of redundant surrounding information, and the target agent may be affected by unimportant neighbors or present overly conservative motion. Thus, we propose Gumbel Social Transformer, in which an Edge Gumbel Selector samples a sparse interaction graph of partially detected pedestrians at each time step. A Node Transformer Encoder and a Masked LSTM encode pedestrian features with sampled sparse graphs to predict trajectories. We demonstrate that our model overcomes potential problems caused by the aforementioned assumptions, and our approach outperforms related works in trajectory prediction benchmarks. Code is available at \url{https://github.com/tedhuang96/gst}.
\end{abstract}
\section{Introduction}\label{sec:Introduction}
\IEEEPARstart{A}{utonomous} mobile robots must comprehensively understand dynamic human environments to safely and smoothly enter our daily lives~\cite{chen2017socially, rudenko2020human}. A human-centered robot should effectively encode motion patterns of surrounding pedestrians from observation, accurately predict their future trajectories, and efficiently plan its own paths for safe and rapid task execution~\cite{ziebart2009planning, du2020online}. Significant progress has been made in understanding human-human interaction and predicting trajectories of multiple pedestrians~\cite{alahi2016social, vemula2018social, gupta2018social, zhang2019sr}, which inspired new contributions in crowd navigation~\cite{chen2019crowd, liu2020decentralized}.

Despite the fruitful results in building socially aware architectures for multi-pedestrian trajectory prediction, previous works usually hold two assumptions which may burden their robotic applications. The first assumption is that positions of all pedestrians are successfully tracked at all times. The second assumption is the target agent (pedestrian or robot) pays attention to all pedestrians in the public scene~\cite{gupta2018social}.

The first assumption defines \emph{fully detected pedestrians} as people who are tracked at every time step during the considered observation and prediction period. This assumption implies that only the fully detected pedestrians are considered for modeling social interaction and predicting trajectories. In contrast, \emph{partially detected pedestrians} comprise both fully detected pedestrians, and people whose positions are tracked for only a proportion of the considered period. Thus, pedestrians who enter the scene later than the beginning of the considered period and who exit earlier than the end are included in partially detected pedestrians. Partially detected pedestrians provide complete pedestrian data, while considering only the fully detected pedestrians results in 40.7\% pedestrians ignored in benchmark datasets. The incomplete pedestrian data caused by the first assumption leads to biased modeling in social interactions. The second assumption requires \emph{full connection} among all pedestrians in the scene, and causes pedestrians that are clearly non-influential to affect motion of the target agent. A workaround for the second assumption would be to restrict the target agent to pay attention to pedestrians in a pre-defined neighborhood and neglect the distant ones~\cite{alahi2016social}. However, the joint influence from too many neighbors in close proximity would still potentially impair feature encoding of the target agent. With redundant concerns on the insignificant surrounding factors, excessively conservative agent behavior may be much like the notorious freezing robot problem~\cite{trautman2010unfreezing}.

We propose Gumbel Social Transformer (GST), which is composed of Edge Gumbel Selector, Node Transformer Encoder, and Masked LSTM. Each component is designed to be capable of processing features of partially detected pedestrians. As for the attention-to-all assumption, we formulate a directed interaction graph to represent the relationship of partially detected pedestrians at each time step. In the interaction graph, a node represents a pedestrian, and a directed edge represents a connection that the node at its tail pays attention to the node at its head. The graph is initialized with full connection which is equivalent as attention to all pedestrians. We apply the Edge Gumbel Selector to prune the edges by following an important constraint: \emph{The target agent can pay attention to at most $n$ pedestrians}. The hyperparameter $n$ is to control the graph sparsity. With the most important relationships preserved between each agent and its $n$ neighbors at each time step, the sparse interaction graphs inferred by the Edge Gumbel Selector are stacked in sequence. The sequence is then fed into the Node Transformer Encoder and the Masked LSTM to spatially and temporally encode features of partially detected pedestrians, and recursively predict their trajectories. 

Our contributions are fourfold: (1) We present a novel architecture to predict trajectories of partially detected pedestrians; (2) We introduce an Edge Gumbel Selector to sample dynamic and sparse interaction graphs of partially detected pedestrians; (3) We demonstrate in multi-agent simulation that our model mitigates the freezing robot problem and diminishes the disturbance from non-influential neighbors on the target agent; and (4) Our model outperforms state-of-the-art approaches on public human trajectory datasets.
\section{Related Work}\label{sec:related}

\textbf{Pedestrian Trajectory Prediction.} Early works have exhaustively investigated hand-engineered features of pedestrian motion~\cite{helbing1995social, van2008reciprocal}. These works perform well in certain cases, but have non-negligible limitations like fixed motion patterns across all pedestrians~\cite{ferrer2014behavior}. Substantial contributions are made to resolve the problems by the integration of socially-aware structures and deep learning methods, including Social LSTM~\cite{alahi2016social}, Generative Adversarial Networks~\cite{gupta2018social}, Self-Attention Mechanism~\cite{vemula2018social, zhang2019sr}, Graph Neural Networks~\cite{huang2019stgat, Mohamed_2020_CVPR}, and Transformer~\cite{yu2020spatio}. However, the assumptions of fully detected pedestrians and global attention over the scene are enforced in many previous works~\cite{vemula2018social, gupta2018social, huang2019stgat, Mohamed_2020_CVPR}. For instance, the binary attention mask in Transformer-based Graph Convolution represents connection between pedestrians, and is set as a fully connected all-one square matrix with the dimension as the amount of fully detected pedestrians in the whole public scene~\cite{yu2020spatio}. Other works constrain the target agent to pay attention within a small neighborhood region, or do not clarify how motion prediction on fully detected pedestrians would be affected by considering partially detected pedestrians~\cite{alahi2016social, zhang2019sr}. In contrast to these works, our work infers a sparse interaction graph among pedestrians in an unsupervised way, and we explicitly study the influence of partially detected pedestrians on trajectory prediction.

\textbf{Graph Structure Learning.} Graph generation has a wide range of applications including causal discovery~\cite{zhu2019causal}, neural architecture search~\cite{xie2018snas}, molecule design~\cite{jin2018junction}, and physical interaction inference~\cite{kipf2018neural}. Traditional approaches are typically hand-crafted for a specific family of graphs~\cite{erdHos1960evolution}, whereas deep learning has recently been harnessed to learn graphs with suitable properties from observation data. One direction for graph generation is to perform sequential prediction on the next node or edge to be added to the graph~\cite{you2018graphrnn}. If the number of graph nodes is fixed, another direction is to generate the adjacency matrix in one shot~\cite{anand2018generative}. Dropout on the adjacency matrix (i.e., dropout on edges) is often used to alleviate over-fitting and over-smoothing~\cite{vaswani2017attention, rong2019dropedge}. Besides regularization, graph sparsity is emphasized in many domains where sparse graph representations are necessary to efficiently learn model parameters~\cite{xie2018snas, kipf2018neural}. Probability distribution of edges are usually assumed independent Bernoulli variables (existence of a single edge)~\cite{franceschi2019learning} or independent categorical variables (type of a single edge)~\cite{kipf2018neural, li2019actional}. In our work, we consider the categorical distribution \textit{over} edges which connect neighbor pedestrian nodes to the same target node. The inferred graph changes dynamically, which is consistent with the dynamic property of pedestrian interactions. 
\section{Method}\label{sec:method}

\subsection{Problem Formulation}\label{sec:method-problem_formulation}

Consider $N$ partially detected pedestrians who appear in a scene during an observation period $t\!\in\!\{1, \ldots, T_{obs}\}$. Their 2D positions are denoted by $x_i^{t}$, $i\!\in\!\{1, \ldots, N\}$. The task is to jointly predict their trajectories $x_i^{t}$ during a following prediction period $t\!\in\!\{T_{obs}+1, \ldots, T_{obs}+T_{pred}\}$. These partially detected pedestrians enter the scene at or later than $t\!=\!1$. They leave the scene at or earlier than $t\!=\!T_{obs}+T_{pred}$. 

We introduce \emph{interaction graphs} to represent motion of partially detected pedestrians. A directed interaction graph $G^{t}\!=\!\left(V^{t}, E^{t}, M^{t}, A^{t}\right)$ describes pedestrian motion at a time step $t$. The set of nodes $V^{t}\!=\!\left\{v_i^t\right\}_{i=1:N}$ corresponds to pedestrian displacement (i.e., velocity). The set of edges $E^{t}\!=\!\left\{e_{ij}^t\right\}_{i,j=1:N}$ corresponds to the relative position from a target pedestrian $i$ to a neighbor~$j$. The node masks $M^{t}\!=\!\left\{m_i^t\right\}_{i=1:N}$ indicate whether the $i$th pedestrian's position is recorded at both $t-1$ and $t$, and the binary-valued adjacency matrix $A^{t}\!=\!\left\{a_{ij}^t\right\}_{i,j=1:N}$ specifies the validity of edges as in Equation~\ref{eq:m-a-v-e-def}, where the edge $e_{ij}^{t}$ is nonexistent whenever either $v_{i}^{t}$ or $v_{j}^{t}$ is invalid. This setting guarantees the full connectivity of an initialized interaction graph $G^t$ by removing the node of a pedestrian, who has not shown up yet or has left the scene, along with all relevant edges. The linear embedding layers and the masks for nodes and edges are applied to respective attributes to obtain high dimensional features, which are still denoted by $v_i^t$ and $e_{ij}^t$ as in Equation~\ref{eq:m-a-v-e-def}.
\vspace{-2pt}
\begin{equation}\label{eq:m-a-v-e-def}
    \begin{aligned}
        m_i^t = \mathds{1}\left\{x_i^{t-1} \textrm{ and } x_i^t \textrm{ are valid} \right\}, \quad a_{ij}^{t} = m_i^t \, m_j^t, \\
        v_i^t = m_i^t \,\,\phi_v\!\left(x_i^t - x_i^{t-1}\right), \quad e_{ij}^{t} = a_{ij}^{t} \,\, \phi_e\!\left(x_j^t - x_i^t\right). \\
    \end{aligned}
\end{equation}

\subsection{Gumbel Social Transformer}
\begin{figure*}[t]
    \centering
    \includegraphics[width=\linewidth]{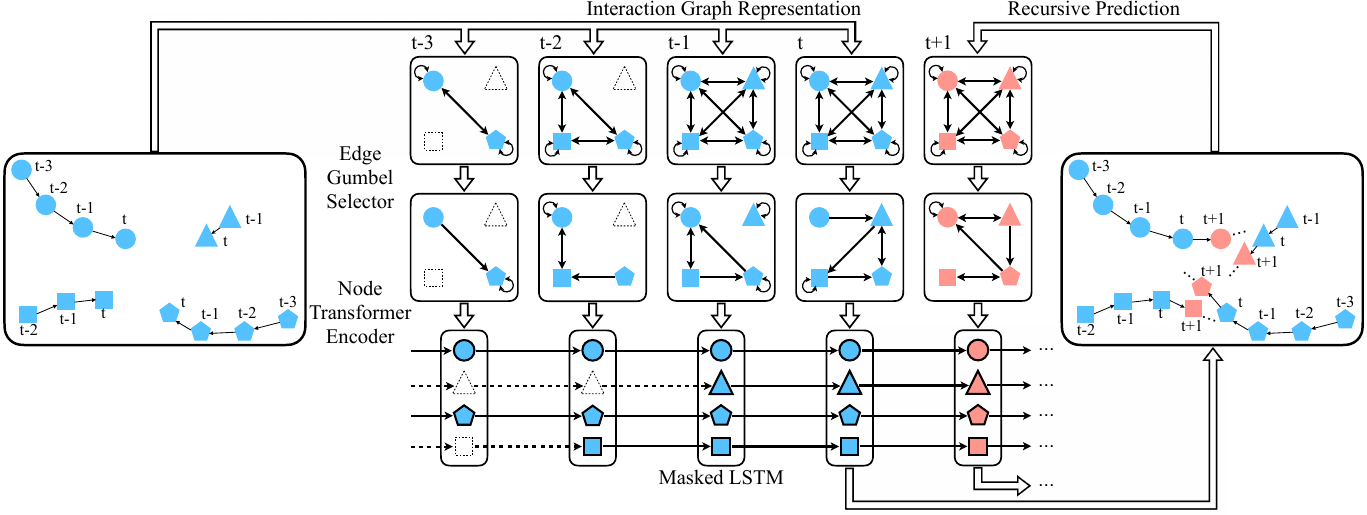}
    \caption{Overview of Gumbel Social Transformer (GST). Observed trajectories (blue) are processed into interaction graph representations at each time step, which are fully connected except for pedestrians not observed at that moment. Under the $n$-neighbor sparsity constraint, Edge Gumbel Selector samples sparse interaction graphs, which are then encoded by Node Transformer Encoder and Masked LSTM. The encoded pedestrian features are used to recursively predict future trajectories (red).}
    \label{fig:gst-architecture}
    \vspace{-20pt}
\end{figure*}

The architecture of Gumbel Social Transformer is illustrated in Fig.~\ref{fig:gst-architecture}. An Edge Gumbel Selector takes as input a combination of node and edge representations from interaction graphs, and samples a sparse interaction graph at each observation time step. A Node Transformer Encoder spatially aggregates node representations of the sampled sparse interaction graphs. The spatially encoded node features are sequentially fed into a Masked LSTM, from which hidden states are used to predict pedestrian positions at the next step. The recursion of feature embedding, edge sampling, node encoding, and node decoding is repeated until the end of the prediction period.

\textbf{Edge Gumbel Selector.} Though the initialized interaction graph $G^t$ includes complete details of the pedestrian motion at time $t$, full connectivity could be redundant, and could even adversely affect the modeling of a target pedestrian $i$'s behavior. We impose the $n$-neighbor sparsity constraint on the interaction graph, which leads to a sparse interaction graph $\Tilde{G}^t\!=\!\left(\Tilde{V}^{t}, \Tilde{E}^{t}, \Tilde{M}^{t}, \Tilde{A}^{t}\right)$. While $\Tilde{V}^{t}, \Tilde{E}^{t}, \Tilde{M}^{t}$ are identical to the counterparts in~$G^t$, the weighted adjacency matrix $\Tilde{A}^t\!=\!\left\{\Tilde{a}_{ij}^t\right\}_{i,j=1:N}\!\in\![0,1]^{N\times N}$ becomes a sparse float-valued matrix to be inferred.

The inference of the adjacency matrix $\Tilde{A}^t$ is formulated as the problem to find the $n$ neighbors who have the most influence on a target pedestrian $i$ at time $t$. We first concatenate neighbor node features, target node features, and edge features to obtain augmented edge features $\hat{e}_{ij}^t$, which represents the pairwise interaction between target $i$ and a neighbor $j$. A neighbor may draw the attention from the target that was originally paid to another neighbor. The relationship of these pairwise interactions $\hat{e}_{ij}^t$ themselves is captured by a multi-head attention (MHA) at the edge level. The number of heads is set as $n$, where each head can be interpreted as one type of the interaction relationship:
\begin{equation*}
\begin{aligned}
\hat{e}_{ij}^{t} &= [v_{j}^{t}\|v_{i}^{t}\|e_{ij}^{t}], \\
\left\{\hat{e}_{ij}^{t,k}\right\}_{j=1:N}^{k=1:n} &= \textrm{MHA}\!\left(\left\{\hat{e}_{ij}^{t}\right\}_{j=1:N}, \textrm{mask}\!=\!\left\{a_{ij}^{t}\right\}_{j=1:N}\right).\\
\end{aligned}
\end{equation*}

A multi-layer perceptron (MLP) maps the aggregated augmented edge features $\hat{e}_{ij}^{t,k}$ to log probabilities $\alpha_{ij}^{t,k}$ of a $N$-dimensional categorical distribution corresponding to the $k$th head. A reparameterization trick named Gumbel Softmax~\cite{jang2017categorical} is used to sample the most important neighbor to the $i$th target from these categorical distributions at each head while preserving differentiability. The samples are drawn from the concrete distribution approximation~\cite{maddison2017concrete} as presented in Equation \ref{eq:gumbel-softmax}, where $\boldsymbol{g} \in \mathbb{R}^{N}$ is a vector with elements sampled from independent and identically distributed random variables with $\textrm{Gumbel(0, 1)}$ distribution.
\begin{equation}\label{eq:gumbel-softmax}
    \begin{aligned}
        \alpha_{ij}^{t,k} &= \textrm{MLP}\!\left(\hat{e}_{ij}^{t,k}\right),\\
        \Tilde{a}_{ij}^{t,k} &= \mathop{\textrm{softmax}}_j\!\left(\left(\alpha_{ij}^{t,k}+\boldsymbol{g}\right)/\tau\right),\\
        \Tilde{a}_{ij}^{t} &= \frac{1}{n}\sum_{k} \Tilde{a}_{ij}^{t,k}.\\
    \end{aligned}
\end{equation}
The temperature hyperparameter $\tau$ in Equation \ref{eq:gumbel-softmax} is annealed to near zero during training, so the approximate samples gradually converge to one-hot samples from the categorical distributions. The entries of the sampled weighted adjacency matrix $\Tilde{a}_{ij}^{t}$'s are the mean of generated samples across the heads. Note the sampling process assures that the set of invalid edges in $\Tilde{E}^t$ is a subset of the set of the removed edges in $\Tilde{A}^t$.

\textbf{Node Transformer Encoder.} Given the sparse interaction graph $\Tilde{G}^t$, the node features are spatially aggregated using an encoder inspired by Transformer-based Graph Convolution (TGConv)~\cite{yu2020spatio}. The Node Transformer Encoder in our case takes the weighted adjacency matrix $\Tilde{A}^{t}$ as the attention mask in TGConv, which is a sparse float matrix, and the inputs of source and target of Transformer are the representations of partially detected pedestrians $\Tilde{V}^t$.
\begin{equation*}
    \left\{\hat{v}^t_i\right\}_{i=1:N} = \textrm{TGConv}\!\left(\textrm{target}\!=\!\Tilde{V}^t\!, \textrm{source}\!=\!\Tilde{V}^t\!, \textrm{mask}\!=\!\Tilde{A}^t\right)
\end{equation*}
\textbf{Masked LSTM.} Pedestrian motion during the observation period is sliced into a stack of interaction graphs $\left\{G^t\right\}^{t=1:T_{obs}}$, which are processed through the Edge Gumbel Selector and the Node Transformer Encoder to obtain spatially encoded node features $\left\{\hat{v}^t_i\right\}_{i=1:N}^{t=1:T_{obs}}$. The node features are sequentially fed into a Masked LSTM to propagate hidden features $h_i^{t}$, which are used to predict pedestrian positions through a linear layer. The node masks $m_i^t$'s are set as one through the prediction period for all pedestrians except for who disappear before or at $T_{obs}$, as we assume they will never come back into the scene. The recurrence is introduced by generating the interaction graph at the next step with the predicted positions.
\begin{equation}\label{eq:pred-output}
    \begin{aligned}
    h_i^{t+1} &= \left(1-m_i^t\right)h_i^{t} + m_i^t \,\textrm{LSTM}\!\left(\hat{v}_i^t, h_i^{t}\right), \\ \hat{x}_i^{t+1} &= \hat{x}_i^{t} + \phi_h\!\left(h_i^{t+1}\right). \\
    \end{aligned}
\end{equation}

\section{Experiments}\label{sec:experiments}

\begin{table*}[hbt!]
\caption{Quantitative performance of all approaches on benchmark datasets. Three metrics Average Offset Error (AOE, unit:~$m$), Final Offset Error (FOE, unit:~$m$), and Negative Log Likelihood (NLL, no unit) on fully detected pedestrians are reported. GST (D) is Gumbel Social Transformer with a deterministic function $\phi_h$ in Equation \ref{eq:pred-output}, while GST (P) is with a probabilistic function $\phi_h$ that outputs Gaussian displacements. GST would be referred to as GST (D) if no additional clarification is given in this work. N/A presented in NLL is due to completely deterministic outputs, on which kernel density estimation cannot be performed.}\label{table:benchmark}
\begin{center}
\setlength\tabcolsep{3pt}
\begin{tabular}{c|c|ccccc|c}
    \toprule
    Metric & Method & ETH & HOTEL & UNIV & ZARA1 & ZARA2 & AVG \\
    \midrule
    \midrule
    \multirow{8}{*}{AOE $\downarrow$} & SLSTM (D) & 2.45$\pm$0.00 & 0.81$\pm$0.00 & 1.24$\pm$0.00 & 2.48$\pm$0.00 & 1.07$\pm$0.00 & 1.61$\pm$0.00 \\
    & STGCN (P) & 2.93$\pm$0.59 & 1.07$\pm$0.21 & 0.76$\pm$0.05 & 0.95$\pm$0.23 & 0.87$\pm$0.14 & 1.32$\pm$0.25 \\
    & STGCN (D) & 2.67$\pm$0.00 & 0.74$\pm$0.00 & 0.64$\pm$0.00 & 0.68$\pm$0.00 & 0.59$\pm$0.00 & 1.06$\pm$0.00 \\
    & SGAN (P) & 1.78$\pm$0.55 & 0.32$\pm$0.08 & 0.62$\pm$0.03 & 0.59$\pm$0.11 & 0.43$\pm$0.10 & 0.75$\pm$0.17 \\
    & STGAT (P) & 1.51$\pm$0.68 & 0.26$\pm$0.09 & 0.57$\pm$0.08 & 0.52$\pm$0.19 & 0.45$\pm$0.16 & 0.66$\pm$0.24 \\
    & Trajectron++ (P) & 1.40$\pm$0.56 & 0.27$\pm$0.17 & \textbf{0.50}$\pm$0.23 & 0.50$\pm$0.23 & 0.33$\pm$0.16 & 0.60$\pm$0.27 \\
    & GST (P) & 1.10$\pm$0.14 & 0.22$\pm$0.02 & 0.64$\pm$0.04 & 0.59$\pm$0.03 & 0.46$\pm$0.03 & 0.60$\pm$0.05 \\
    & GST (D) & \textbf{0.96}$\pm$0.20 & \textbf{0.21}$\pm$0.02 & \textbf{0.50}$\pm$0.01 & \textbf{0.40}$\pm$0.00 & \textbf{0.32}$\pm$0.02 & \textbf{0.48}$\pm$0.05 \\
    
    \midrule
    \multirow{8}{*}{FOE $\downarrow$} & SLSTM (D) & 4.20$\pm$0.00 & 1.46$\pm$0.00 & 2.20$\pm$0.00 & 4.49$\pm$0.00 & 1.93$\pm$0.00 & 2.86$\pm$0.00 \\
    & STGCN (P) & 4.98$\pm$0.89 & 1.54$\pm$0.29 & 1.39$\pm$0.09 & 1.48$\pm$0.30 & 1.31$\pm$0.20 & 2.14$\pm$0.35 \\
    & STGCN (D) & 4.83$\pm$0.00 & 1.23$\pm$0.00 & 1.26$\pm$0.00 & 1.28$\pm$0.00 & 1.06$\pm$0.00 & 1.93$\pm$0.00 \\
    & SGAN (P) & 3.60$\pm$1.33 & 0.60$\pm$0.19 & 1.31$\pm$0.06 & 1.28$\pm$0.26 & 0.94$\pm$0.21 & 1.55$\pm$0.41 \\
    & STGAT (P) & 3.01$\pm$1.41 & 0.48$\pm$0.21 & 1.23$\pm$0.18 & 1.15$\pm$0.46 & 1.02$\pm$0.42 & 1.38$\pm$0.54 \\
    & Trajectron++ (P) & 2.96$\pm$1.27 & 0.53$\pm$0.37 & 1.27$\pm$0.61 & 1.14$\pm$0.58 & 0.80$\pm$0.43 & 1.34$\pm$0.65 \\
    & GST (P) & 2.22$\pm$0.33 & \textbf{0.37}$\pm$0.04 & 1.32$\pm$0.09 & 1.15$\pm$0.06 & 0.95$\pm$0.07 & 1.20$\pm$0.12 \\
    & GST (D) & \textbf{2.09}$\pm$0.47 & 0.38$\pm$0.04 & \textbf{1.08}$\pm$0.02 & \textbf{0.86}$\pm$0.00 & \textbf{0.70}$\pm$0.04 & \textbf{1.02}$\pm$0.11 \\
    \midrule
    \multirow{6}{*}{NLL $\downarrow$} & SLSTM (D) & N/A & N/A & N/A & N/A & N/A & N/A \\
    & STGCN (P) & 10.07 & 3.57 & 3.06 & 4.20 & 2.13 & 4.61\\
    & STGCN (D) & N/A & N/A & N/A & N/A & N/A & N/A \\
    & SGAN (P) & 8.11 & 11.99 & 13.56 & 9.20 & 2.61 & 9.09 \\
    & STGAT (P) & 3.82 & 1.84 & 4.68 & 2.15 & -0.16 & 2.47 \\
    & Trajectron++ (P) & 1.76 & -1.29 & -0.75 & -0.41 & -1.83 & -0.50 \\
    & GST (P) & \textbf{-0.44} & \textbf{-1.79} & \textbf{-1.98} & \textbf{-2.08} & \textbf{-3.61} & \textbf{-1.98} \\
    & GST (D) & 8.02 & 9.29 & 16.00 & N/A & 9.10 & N/A \\
    \bottomrule
\end{tabular}
\end{center}
\vspace{-20pt}
\end{table*}

We use two publicly available trajectory datasets for benchmarking: ETH~\cite{pellegrini2009you} and UCY~\cite{lerner2007crowds}. ETH is composed of two scenes ETH and HOTEL, and UCY is composed of three scenes UNIV, ZARA1, and ZARA2. The frame rate is 2.5 frames per second across the scenes. The task is to predict the trajectories in the next 4.8 seconds ($T_{pred}\!=\!12$) given the positions tracked in the last 3.2 seconds ($T_{obs}\!=\!8$). There are in average 40.7\% of all detected pedestrians who are partially but not fully detected among these datasets. The trajectory samples in each scene are split into the training set (80\%) and the test set (20\%), and models are independently trained for each scene. Additionally, we conduct comparative study on our models with various configurations. A multi-agent simulation is performed with models trained in the comparative study, to evaluate their capabilities to generate trustworthy future trajectories in nontrivial social interaction scenarios.

\vspace{-5pt}
\subsection{Implementation Details}
The embedding dimension of nodes is 32 and of edges is 64. The dimension of hidden states in LSTM is 32. The temperature $\tau$ of the Edge Gumbel Selector is annealed linearly from 0.5 to 0.03 through training. We empirically find a higher initial temperature is likely to result in numerical instability during training. The Node Transformer Encoder has 3 Transformer Encoder layers with 8 attention heads, and a feed-forward dimension of 128. The Adam optimizer~\cite{kingma2015adam} is used to minimize the mean square error loss of prediction on trajectories of partially detected pedestrians, with an initial learning rate of 0.001. The model is trained for 200 epochs. Trajectory samples are randomly rotated during the training process for data augmentation. A \emph{ghost agent} with zero-valued features is added in sparse configurations for promoting sparsity, which is inspired by the ghost link in~\cite{li2019actional}.

\begin{figure*}[t!]
    \centering
    \includegraphics[width=\linewidth]{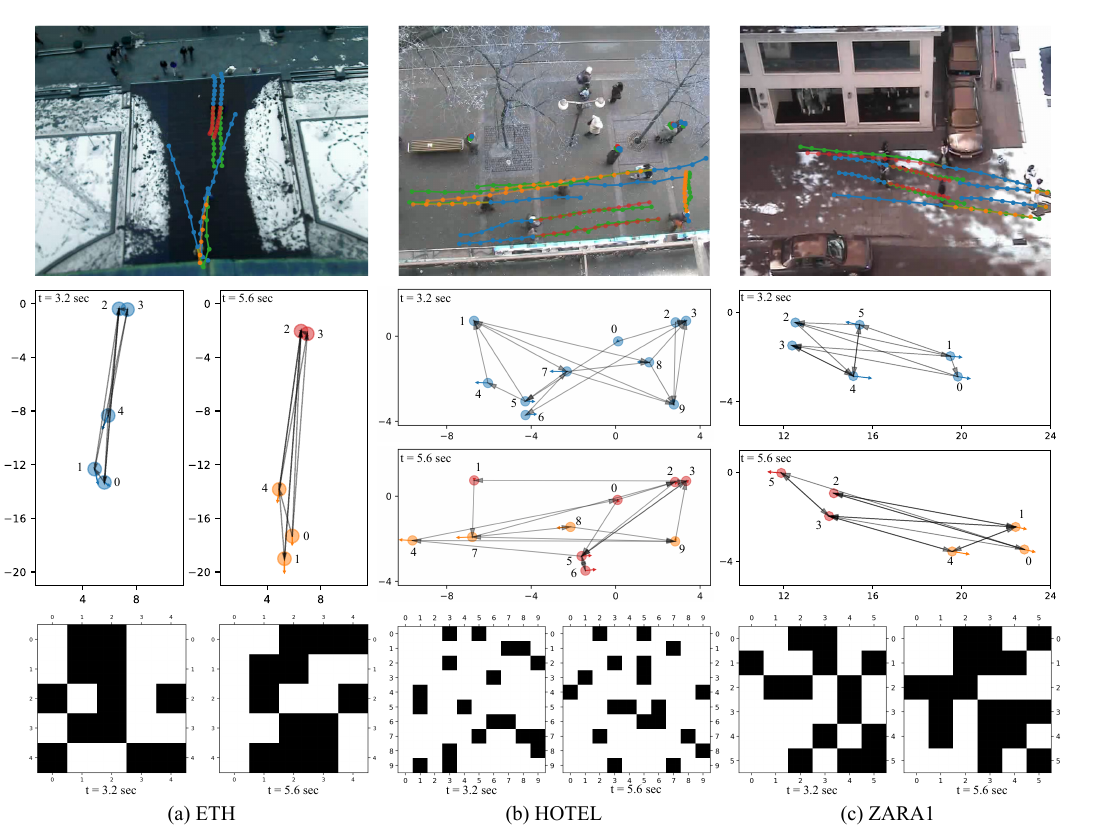}
    \caption{Top row visualizes trajectory prediction on benchmark datasets. Blue denotes observation, green denotes ground truth, red denotes prediction on fully detected pedestrians, and orange denotes prediction on the other partially detected pedestrians. Middle row demonstrates the sparse interaction graph inference at the last observed time step, and at the middle of the prediction period. Colored arrows indicate pedestrian velocities. Gray arrows represent directed edges of the interaction graph, where the target pedestrian node at the tail pays attention to the neighbor node at the head. Bottom row shows the adjacency matrices corresponding to the inferred graphs in the middle row. The black entry at the $i$th row, the $j$th column indicates the $i$th target pays attention to the $j$th neighbor.}
    \label{fig:benchmark}
    \vspace{-20pt}
\end{figure*}
\vspace{-3pt}
\subsection{Benchmark Evaluation}
\textbf{Baselines.} Our model is compared against these existing methods: (1) Social LSTM (SLSTM) is a LSTM integrated with a social pooling layer that outputs deterministic trajectories~\cite{alahi2016social}; (2) Social STGCNN (STGCN) is a spatial graph convolution network concatenated with a temporal convolution network that generates probabilistic outputs~\cite{Mohamed_2020_CVPR}; (3) The variant of STGCN which generates deterministic prediction; (4) Social GAN (SGAN) has a LSTM-based encoder-decoder architecture with a socially aware global pooling layer, and is trained using Generative Adversarial Networks for multi-modal trajectory prediction~\cite{gupta2018social}; (5) Spatial-Temporal GAT (STGAT) applies graph attention networks to model crowd interaction, and uses different LSTMs for temporal encoding of single pedestrians and of spatially encoded interaction~\cite{huang2019stgat}; (6) Trajectron++ is a conditional variational autoencoder which encodes agent interaction through attention mechanism and semantic map through convolutional neural networks~\cite{salzmann2020trajectron++}.

\textbf{Metrics.} Offset Error is defined as the distance between the target pedestrian's ground truth position and the position predicted by the model at one time step~\cite{alahi2016social, huang2021long}. Average Offset Error (AOE) is the average of Offset Errors throughout the prediction period, and Final Offset Error (FOE) is the Offset Error at the last prediction step $T_{obs}+T_{pred}$. For evaluation of probabilistic methods, 20 trajectories are predicted, over which the mean and the standard deviation of AOEs and FOEs are reported. To evaluate trajectory sample distribution, we calculate Negative Log Likelihood (NLL) of the ground truth trajectory under the distribution from kernel density estimation~\cite{salzmann2020trajectron++}. 

\textbf{Results.} The quantitative results are presented in Table~\ref{table:benchmark}. Note that our model GST is trained with partially detected pedestrians, while other baselines have to be trained with only fully detected pedestrians. Nevertheless, all metrics are only reported on fully detected pedestrians for fair comparison across the methods. Our model GST (D),which is GST with a deterministic output function $\phi_h$ in Equation \ref{eq:pred-output}, exceeds mean AOE/FOE performance of other state-of-the-art approaches on most datasets. However, large NLL of GST (D) indicates relative uni-modality of the prediction samples in contrast to probabilistic baselines, because the stochasticity of GST (D) is only from sampling of the sparse interaction graphs. A more diverse trajectory sample distribution can be achieved by applying a Gaussian displacement output function, which is presented as GST (P) in Table \ref{table:benchmark}.

\textbf{Visualization.} Trajectory prediction and inferred sparse interaction graphs in different scenes are visualized in Fig.~\ref{fig:benchmark}. Our model is able to predict trajectories of all partially detected pedestrians, and also model their interaction by inferring sparse interaction graphs. We visualize these graphs at $t\!=\!3.2\,\textrm{sec}$ and $t\!=\!5.6\,\textrm{sec}$ in each scene, where the former is the last observed time step, and the latter is at the middle of the prediction period. The graph structure varies at different times, indicating the evolution of the relationship between pedestrians due to the change of their positions and velocities. For example, in the HOTEL scene, the 5th and the 6th pedestrians were considered not interactive at $t\!=\!3.2\,\textrm{sec}$, while the mutual attention is added later at $t\!=\!5.6\,\textrm{sec}$ during the recursive prediction. In contrast, there were initially bi-directed edges between the 4th and the 5th pedestrians in the ZARA1 scene at $t\!=\!3.2\,\textrm{sec}$. However, they walked away from each other, and both edges are removed at $t\!=\!5.6\,\textrm{sec}$.

\begin{figure*}[t!]
    \centering
    \includegraphics[width=\linewidth]{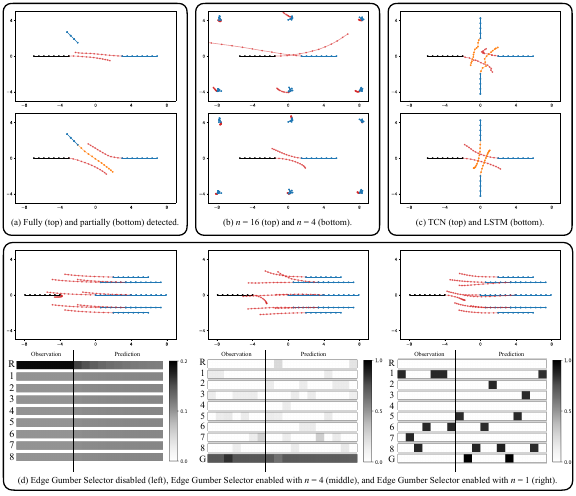}
    \caption{Comparison of multi-agent simulation results in different human-robot interaction scenarios: (a) Human agents enter the robot agent's field of view at different times; (b) Some human agents walk aimlessly in the scene; (c) robot meets humans at intersection; (d) One robot agent encounters a group of human agents. Black and blue represents the observation on robot and human agents. Red indicates prediction on the robot and fully detected humans, and orange indicates prediction on the other partially detected humans. In time-series robot attention plots of (d), R denotes robot, 1-8 denote humans, and G denotes a ghost agent with zero-valued features for encouraging sparsity.}
    \label{fig:comparative}
    \vspace{-20pt}
\end{figure*}

\vspace{-7pt}
\subsection{Comparative Study}

The effect that each component of Gumbel Social Transformer has on the performance of trajectory prediction is assessed by extensive comparative experiments. Besides ETH and UCY, crowd datasets CFF, LCAS, WILDTRACK and SYNTH are adopted from Trajnet++ to investigate how the performance of different configurations varies across datasets with different crowd densities \cite{alahi2014socially, sun20183dof, chavdarova2018wildtrack, kothari2021human}. Moreover, the trained models are directly applied in multi-agent simulation for generating future motion of robot and human agents. The quality of the generated motions is examined in common yet nontrivial human-robot interaction scenarios. The simulation is similar to the setup in~\cite{chen2019crowd, liu2020decentralized}, where holonomic kinematics are used for both robot and human agents, and the displacements of the generated trajectories are action inputs to each agent.

\begin{figure}[t!]
    \centering
    \includegraphics[width=\linewidth]{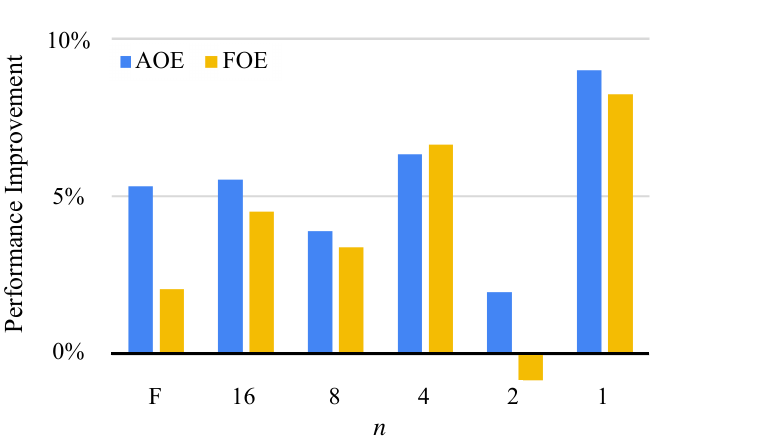}
    \caption{Improvement of trajectory prediction performance by changing from fully detected pedestrians to partially detected pedestrians for different sparse configurations. F denotes full connection.}
    \label{fig:partial_vs_full}
    \vspace{-15pt}
\end{figure}

\textbf{Partially or fully detected Pedestrians.} Fig. \ref{fig:partial_vs_full} reports that under most sparsity configurations, trajectory prediction is improved among all datasets by considering partially detected pedestrians. We reason that trajectories of partially detected pedestrians create a complete picture of pedestrian interaction during the observation period, and thus provides an unbiased input for encoding socially aware pedestrian features. The importance of partially detected pedestrians is illustrated in Fig.~\ref{fig:comparative}~(a), where a robot agent and a fully detected human agent walk against each other. The second human agent appears 1.6 seconds (4 time steps) later and then starts approaching other agents. The top of Fig.~\ref{fig:comparative}~(a) shows the case when both the robot and the fully detected human ignore the partially detected one. In contrast, the bottom of Fig.~\ref{fig:comparative}~(a) illustrates that both agents deviate from the original path to dodge the partially detected human.

\begin{figure}[t!]
    \centering
    \includegraphics[width=\linewidth]{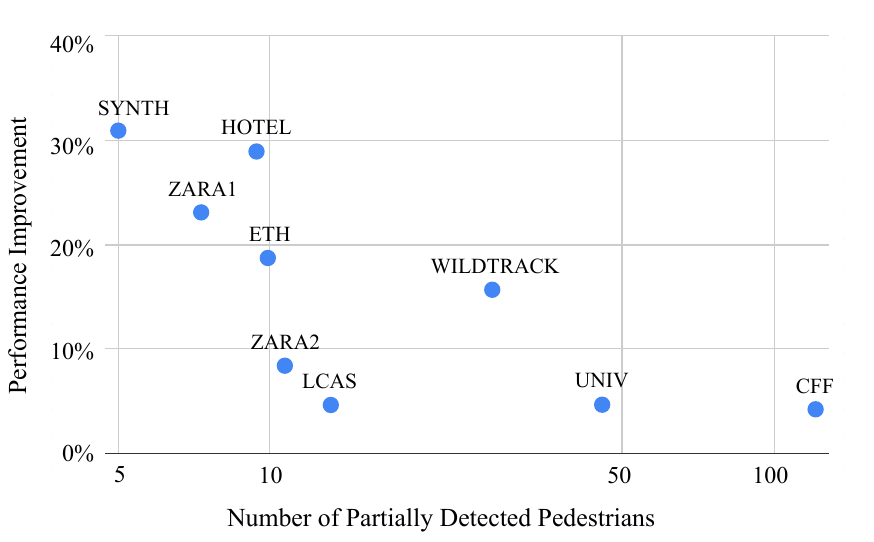}
    \caption{Improvement of trajectory prediction performance by changing from fully connected configuration to the best sparse configuration across datasets.}
    \label{fig:best_vs_baseline}
    \vspace{-15pt}
\end{figure}

\begin{figure}[t!]
    \centering
    \includegraphics[width=\linewidth]{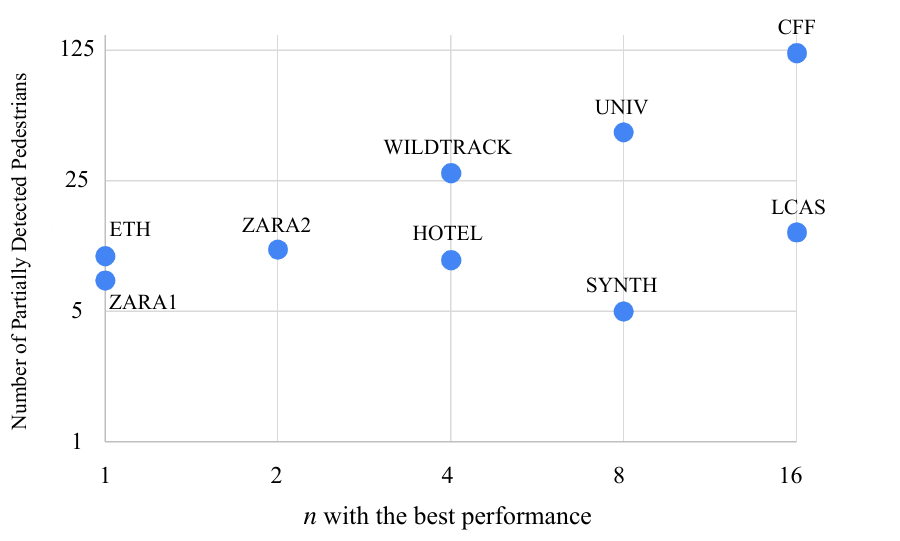}
    \caption{The sparsity hyperparameter $n$ with the best prediction performance across datasets.}
    \label{fig:best_n}
    \vspace{-20pt}
\end{figure}

\textbf{Sparse or Fully Connected Interaction Graph.} The improvement of trajectory prediction by replacing the fully connected interaction graph with the best sparse configuration is demonstrated for each dataset in Fig.~\ref{fig:best_vs_baseline}. We observe a trend that when the average number of partially detected pedestrians is larger in a scene, the improvement is less significant. This phenomenon may be attributed to the fact as presented in Fig.~\ref{fig:best_n}, where the best hyperparameter $n$ is likely to be larger if the considered dataset has a higher crowd density. A larger $n$ indicates more connectivity between a target and neighbors, and thus a narrower gap between the corresponding sparse configuration and the fully connected counterpart.

Sparsity is critical to address the freezing agent problem caused by over-smoothing. As shown in Fig.~\ref{fig:comparative}~(d), a robot agent is moving right and encounters a human crowd of eight moving left. The left of Fig.~\ref{fig:comparative}~(d) presents the results when Edge Gumbel Selector is disabled. The robot agent turned around in order to dodge the crowd. The prediction on robot motion is too conservative to match pedestrian social norms. The prediction of crowd motion which responds to the conservative robot motion would also be erroneous, and can affect the down-stream motion planning pipeline. The time-series attention visualization in Fig.~\ref{fig:comparative}~(d) left indicates the robot paid more attention to itself during the observation period, but then the attention is gradually distributed to human neighbors. We reason that the features of the crowd overwhelmed the robot features in the weighted sum step of self-attention. This smoothing effect is passed with the hidden states and leads to the U-turn behavior. This unnatural turnaround motion is effectively alleviated by enabling Edge Gumbel selector to introduce sparsity, as shown in middle and right of Fig.~\ref{fig:comparative}~(d). Note in Fig.~\ref{fig:comparative}~(d) left, the same attention across human agents and their identical predicted trajectories imply symmetry of human agents in the fully connected interaction graph.

As shown in Fig. \ref{fig:best_n}, the sparsity constraint needs to be tradeoff conditioned on the crowd density of different scenes. It is important to control sparsity level and help target agent concentrate on interaction with critical neighbors. Fig.~\ref{fig:comparative}~(b) illustrates a scene where six human agents are moving randomly near the boundaries, and an important human agent is running into the robot. The top of Fig.~\ref{fig:comparative}~(b) shows the results with $n = 16$, where the robot and the interactive human exhibit exaggerated motion. However, the robot agent naturally avoids collision with the close human neighbor at the bottom of Fig.~\ref{fig:comparative}~(b), where $n$ is set as $4$. This indicates when the interaction graph is almost fully connected, the target agent is easy to be affected by the connected neighbors, even the ones who are clearly non-influential.

\textbf{Recursive or Readout Prediction.} We analyze the effect of recursive trajectory prediction by comparison between Masked LSTM and temporal convolution network (TCN), which has been applied to encode temporal relationship and make sequential prediction~\cite{Mohamed_2020_CVPR, yan2018spatial}.  While quantitative performances are found comparable, we see in Fig. \ref{fig:comparative} (c) the trajectories generated from TCN are less smooth than those from Masked LSTM. The intuition is that TCN implements a multi-layer perceptron to expand temporally encoded features from the sequence length of $T_{obs}$ to $T_{pred}$, and generates distinctive and discontinuous features for prediction of consecutive displacements. In contrast, Masked LSTM encodes the features in time sequence with shared weights, so the temporally encoded features which will be mapped to the predicted displacements keep the continuity, and produce smoother and more natural future trajectories.
\section{Conclusions and Future Work}\label{sec:conclusions}
We identify two common assumptions of existing pedestrian trajectory prediction approaches: pedestrian positions are always successfully tracked, and the target agent pays attention to all pedestrians in the detected range. These assumptions can cause issues of the deployment of trajectory prediction algorithms in real world robot applications. We present Gumbel Social Transformer to overcome these issues. Our model architecture is designed to encode features of partially detected pedestrians, and thus provides a complete input for unbiased modeling on pedestrian interaction. We propose Edge Gumbel Selector, which is an unsupervised method that infers a sequence of sparse interaction graphs to summarize the evolving relationship among pedestrians. We demonstrate the the introduction of sparsity to modeling multi-agent interaction effectively alleviates the freezing robot problem, and minimizes the influence on generating target agent's motion from unimportant neighbors. 

However, we also observe that the performance of a sparsity configuration is dependent on scene properties such as average number of partially detected pedestrians. A fixed sparse hyperparameter would constrain generalization across the scenes. As future work, we will extend our approach with learnable sparsity to handle scenes with varying crowd densities. To finetune our trajectory prediction model for navigation applications, a trajectory dataset of human-robot interaction will be collected with a bird's eye view camera similar to~\cite{ziebart2009planning}. As for deployment on a generic camera-on-robot setup~\cite{liu2020decentralized}, we will explore occlusion inference approaches~\cite{itkina2021multi} to attempt to minimize the influence of long occlusion periods on the trajectory prediction performance.

\bibliographystyle{IEEEtran}
\bibliography{bib}

\end{document}